\documentclass{Odyssey2026}

 \interspeechcameraready

\title{FiLM-Based Speaker Conditioning of a SpeechLLM \\for Pathological Speech Recognition}

\author[affiliation={1,2}]{Fernando}{López}
\author[affiliation={3}]{Santosh}{Kesiraju}
\author[affiliation={1}]{Jordi}{Luque}

\affiliation{Scientific Research}{Telefónica Innovación Digital}{Spain}
\affiliation{AUDIAS}{Universidad Autónoma de Madrid}{Spain}
\affiliation{Speech@FIT}{Brno University of Technology}{Czech Republic}
\email{fernando.lopez@telefonica.com}

\usepackage{comment}
\usepackage{soul}
\usepackage{multirow}
\usepackage{amsmath}
\usepackage{amssymb}
\begin{document}

\maketitle

\begin{abstract}
    

    Automatic speech recognition (ASR) has advanced remarkably for standard speech; however, pathological speech from neurological conditions remains a significant challenge. We investigate speaker conditioning via Feature-wise Linear Modulation (FiLM), injecting x-vector-derived information into each transformer layer of a frozen ASR encoder to adapt internal representations to individual pathological speakers without modifying base model weights. We benchmark this for the ASR task against standard and parameter-efficient fine-tuning baselines, complemented by post-processing, on Spanish and English pathological speech. Additionally, we evaluate if the adapted model preserves the ability to answer speech-related questions. Results show that speaker-conditioned ASR is competitive with established adaptation strategies while retaining performance on non-conditioned speech.

\end{abstract}

\section{Introduction}
\label{sec:intro}

Automatic speech recognition (ASR) has improved markedly in recent years, with large pretrained models achieving low word error rates on standard benchmarks \cite{ahlawat2025automatic}. Yet these systems continue to struggle with speech produced by individuals with neuromotor disorders such as amyotrophic lateral sclerosis (ALS) or Parkinson's disease (PD). These conditions frequently cause dysarthria, a motor speech disorder that impairs articulatory clarity, alters prosody, and produces high inter-speaker variability \cite{rudzicz2012torgo, kim2008dysarthric}, creating an acoustic mismatch with the normative speech on which modern ASR systems are trained.

Domain adaptation is a natural response, and the Interspeech Speech Accessibility Project (SAP) challenge \cite{hasegawainterspeech} targets this problem. Top-performing systems of the challenge demonstrate that standard and parameter-efficient fine-tuning methods, such as Low-Rank Adaptation (LoRA) applied to pretrained models like \texttt{whisper-large-v3}, combined with pre- and post-processing techniques, are effective when sufficient in-domain data is available. However, pathological speech corpora remain scarce and limited in speaker diversity, even for English, despite resources such as UA-Speech \cite{kim2008dysarthric}, TORGO \cite{rudzicz2012torgo}, and the SAP Challenge. The situation is more acute for other languages, such as Spanish, where efforts like the GITA corpus \cite{orozco2014new}, and NeuroVoz \cite{mendes2024neurovoz, mendes_laureano_2024_10777657} are valuable but make data-hungry fine-tuning strategies fragile and prone to overfitting on the few available speakers.

Speaker-level conditioning is a complementary approach, injecting speaker-specific information to help the model adapt to individual acoustic profiles. Prior work has explored x-vector or fMLLR embeddings via bottleneck adapters \cite{baskar2022speaker}, severity-informed learned hidden unit contributions (LHUC) that apply speaker-dependent scaling to hidden activations \cite{geng2023use}, and x-vector projections concatenated into encoder outputs to condition the ASR decoder via cross-attention \cite{wagner2025personalized}. These methods show promise but share a common limitation: they modify pretrained weights, risking degradation on normative speech. This concern is growing as ASR systems evolve toward models that combine audio encoders with large language models (LLMs), speechLLMs \cite{shi2026qwen3}. These models support tasks beyond transcription \cite{ding2025kimi, goel2025audioflamingo3advancing, liu2025voxtral}, making weight preservation and generalization increasingly critical.

To address this, we investigate an adaptation strategy that leaves all base model weights untouched. We condition a frozen encoder with speaker information via Feature-wise Linear Modulation (FiLM) \cite{perez2018film}. Specifically, we inject speaker information after every transformer layer to modulate internal representations through linear transformations. We evaluate this approach on NeuroVoz (Spanish) and TORGO (English) under low-resource conditions. Our contributions are:

\begin{itemize}
    \item We propose a FiLM-based speaker-conditioned adaptation strategy for pathological ASR\footnote{Code: \url{https://github.com/ferugit/film-spk-asr}}. It freezes all base model weights and preserves all base knowledge for healthy speech.
    \item We provide a systematic comparison of four adaptation strategies: full fine-tuning, encoder-only fine-tuning, LoRA, and speaker conditioning. We analyze their trade-offs in the low-resource pathological speech domain.
    \item We evaluate the adapted models within the multiple-choice question answering (MCQA) paradigm. We build speech questions from NeuroVoz and TORGO metadata to assess whether adaptation gains on ASR come at the cost of broader audio paralinguistic question-answering abilities, comparing all approaches against the unmodified base model.
\end{itemize}

\section{Speaker-Conditioned ASR via FiLM Modulation}
\label{sec:film-conditioning}
We condition the ASR encoder of a frozen SpeechLLM on pathological speech by injecting speaker-derived information after every transformer layer. Speaker information is obtained from FiLM layers driven by x-vector speaker embeddings. All pretrained weights of the SpeechLLM remain frozen in our approach. We just adapt the speaker embedding extractor and the FiLM layers.

\subsection{Speaker Embedding Extraction}
\label{sec:xvector}

Given a raw audio waveform $\mathbf{x} \in\mathbb{R}^{T}$, we extract a fixed-dimensional speaker embedding (x-vector) using a pretrained model $f_{\mathrm{spk}}$:
\begin{equation}
  \mathbf{z} = f_{\mathrm{spk}}(\mathbf{x}) \in \mathbb{R}^{d_{\mathrm{spk}}}.
  \label{eq:xvector}
\end{equation}
The speaker embedding model processes raw audio by extracting filter-bank features and projecting them into a $d_{\mathrm{spk}}$-dimensional speaker space. To prevent the encoder from learning speaker cues from healthy speakers, we zero out the embedding for normative utterances. Let $m_i\in\{0,1\}$ be a binary flag indicating whether the $i$-th sample is normative ($m_i=1$ for healthy, $m_i=0$ for pathological). The effective speaker embedding is then:
\begin{equation}
  \tilde{\mathbf{z}}_i = (1 - m_i)\,\mathbf{z}_i.
  \label{eq:normative-mask}
\end{equation}
For normative speakers, $\tilde{\mathbf{z}}_i = \mathbf{0}$, which, combined with the identity initialization described below, ensures that FiLM acts as an identity transformation on healthy speech.

\subsection{FiLM Parameter Generation}
\label{sec:film-generator}

For each encoder layer $\ell \in \{1,\dots,L\}$, a dedicated two-layer multi-layer perceptron (MLP) is the FiLM generator $\mathcal{G}_\ell$) that maps the speaker embedding to affine modulation parameters:
\begin{equation}
  [\,\boldsymbol{\gamma}_\ell\,;\,\boldsymbol{\beta}_\ell\,]
  = \mathcal{G}_\ell(\tilde{\mathbf{z}}),
  \qquad
  \boldsymbol{\gamma}_\ell,\,\boldsymbol{\beta}_\ell \in \mathbb{R}^{d},
  \label{eq:film-mlp}
\end{equation}
where $d$ is the encoder model dimension. The generator first applies layer normalization~(LN) to the input embedding before the MLP:
\begin{equation}
\mathcal{G}_\ell(\tilde{\mathbf{z}})
=
\mathbf{W}_\ell^{2}\,
\operatorname{ReLU}(
\mathbf{W}_\ell^{1}\,\operatorname{LN}(\tilde{\mathbf{z}})
+ \mathbf{b}_\ell^{1})
+ \mathbf{b}_\ell^{2},
\label{eq:film-mlp-internal}
\end{equation}
where $\mathbf{W}_\ell^{1} \in \mathbb{R}^{d_h \times d_{\mathrm{spk}}}$,
$\mathbf{W}_\ell^{2} \in \mathbb{R}^{2d \times d_h}$, and $d_h$ is a hidden
dimension smaller than $d$. LN removes dependency from embedding magnitude, removing scale differences introduced by varying utterance length, loudness, or channel conditions.

In addition to the affine parameters, each generator produces a scalar gate
$\alpha_\ell \in [0,1]$ via a lightweight MLP:
\begin{equation}
\alpha_\ell = \sigma\bigl(\mathcal{G}_\ell^{g}(\tilde{\mathbf{z}})\bigr),
\label{eq:gate}
\end{equation}
where $\sigma$ is the sigmoid function and $\mathcal{G}_\ell^{g}$ shares the same structure as $\mathcal{G}_\ell$ (normalization, linear, ReLU, linear) with a hidden dimension of $d_h/2$. The gate controls the degree of speaker conditioning per layer, interpolating continuously between no modulation ($\alpha_\ell = 0$) and full modulation ($\alpha_\ell = 1$).

\subsection{Feature-wise Linear Modulation}
\label{sec:film-apply}

Let $\mathbf{H}_\ell \in \mathbb{R}^{N \times d}$ denote the output of the $\ell$-th transformer layer of the ASR encoder, where $N$ is the sequence length. The gated FiLM modulation is applied as a learned residual:
\begin{equation}
  \widetilde{\mathbf{H}}_\ell
  =
  \mathbf{H}_\ell
  +
  \alpha_\ell\,\bigl(
    (\boldsymbol{\gamma}_\ell - \mathbf{1}) \odot \mathbf{H}_\ell
    + \boldsymbol{\beta}_\ell
  \bigr),
  \label{eq:film-residual}
\end{equation}
where $\odot$ denotes the Hadamard product, and $\boldsymbol{\gamma}_\ell$, $\boldsymbol{\beta}_\ell$ are broadcast along the time dimension $N$.

\begin{figure}[h]
  \centering
  \includegraphics[width=1.0\columnwidth]{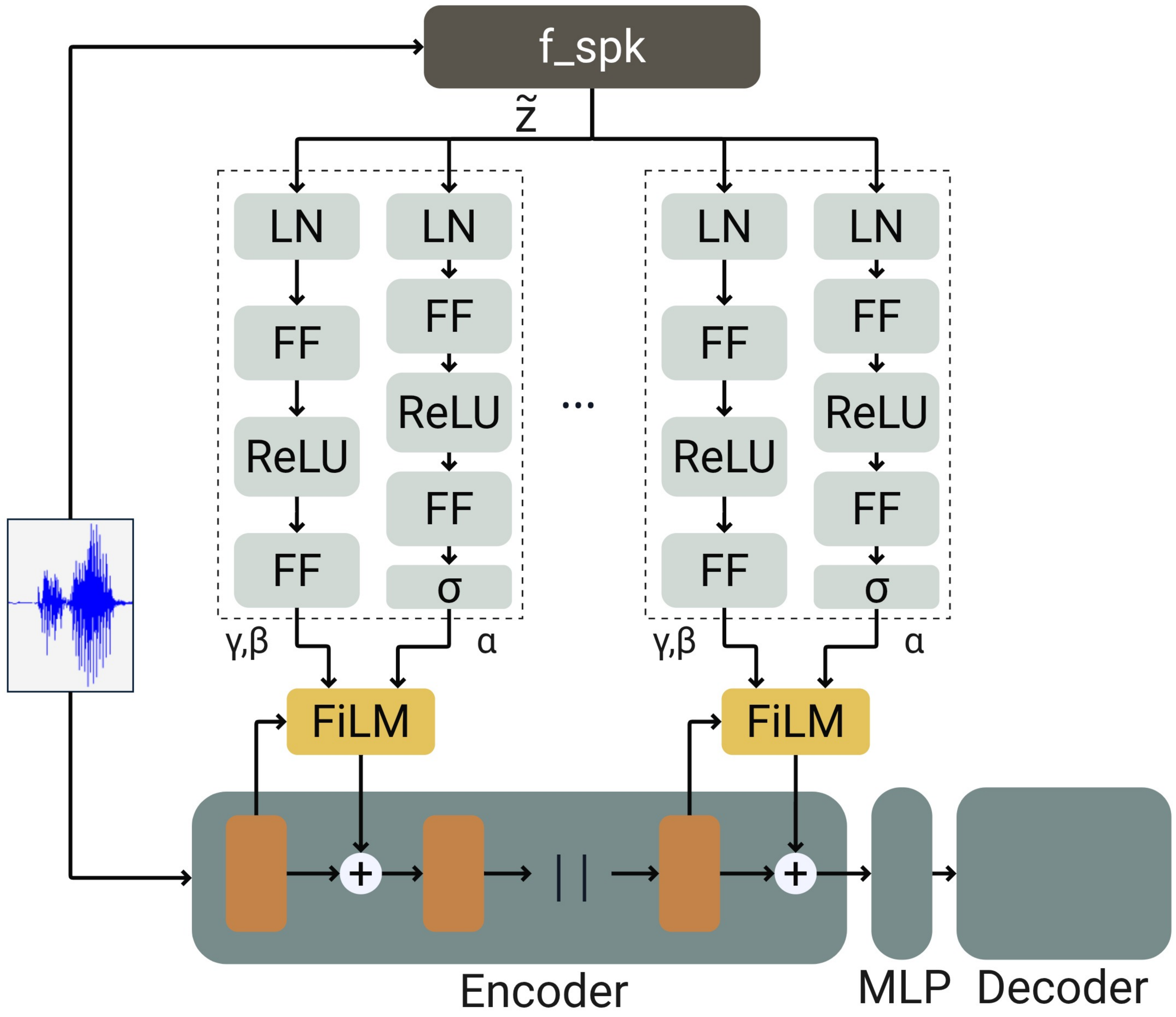}
  \caption{Proposed FiLM conditioning architecture. The raw audio is processed by the speaker encoder $f_{\mathrm{spk}}$ and the ASR encoder. At each transformer layer, the masked x-vector $\tilde{\mathbf{z}}$ is passed through a per-layer FiLM generator that produces affine parameters $(\boldsymbol{\gamma}_\ell, \boldsymbol{\beta}_\ell)$ and a scalar gate $\alpha_\ell$, which jointly modulate the layer output. The MLP connector bridges the conditioned encoder output to the LLM-based decoder.}
  \label{fig:method-scheme}
\end{figure}

Figure~\ref{fig:method-scheme} illustrates the proposed architecture. The conditioned representations $\widetilde{\mathbf{H}}_\ell$ then flow through the remainder of the network: the connector and the LLM-based decoder, both of which remain frozen.

\section{Experimental Setup}
\label{sec:experimental-setup}

We compare our method against adaptation strategies explored in the SAP challenge \cite{hasegawainterspeech}, namely standard fine-tuning and parameter-efficient fine-tuning, further complemented by a text post-processing stage. Each model is first fine-tuned on the pathological speech corpora for the ASR task and then evaluated on ASR performance. Additionally, we assess whether adaptation comes at the cost of losing broader speech understanding by constructing an MCQA benchmark from speaker metadata in the evaluation sets.

\subsection{ASR Model}
\label{sec:asr-model}

We experiment with \textbf{Voxtral-Mini} \cite{liu2025voxtral}, an open-weight speech-LLM with 4.7B parameters. It comprises a fine-tuned Whisper-large-v3 encoder connected through an MLP to a Ministral-3B LLM backbone \cite{liu2026ministral3}, supporting audio-conditioned instruction-following across 8 languages beyond transcription. We use the \texttt{Voxtral-Mini-3B-2507} checkpoint.

\subsection{Speaker Embedding Model}
\label{sec:speaker-model}
Since our experiments span Spanish and English, we adopt a multilingual pretrained Simple Attention Module ResNet-34 (\textbf{SiAmResNet34})\footnote{We use the pretrained model from the WeSpeaker toolkit: \url{https://github.com/wenet-e2e/wespeaker}}, totaling 25.2M parameters. The model was pretrained on VoxBlink2
\cite{lin2024voxblink2} and fine-tuned on VoxCeleb2 \cite{chung2018voxceleb2}. It takes raw waveforms as input, computes 80-dimensional filter-bank features, and outputs $d_{\mathrm{spk}}$-dimensional x-vector speaker embeddings.

\subsection{Datasets}
\label{sec:databases}

We use two pathological speech corpora, both as adaptation training and evaluation data.

\textbf{TORGO} \cite{rudzicz2012torgo} is an English dysarthric speech corpus of speakers with cerebral palsy, covering diadochokinetic tasks, sustained phonation, isolated words, and constrained and unconstrained sentences. We retain the isolated-word and constrained-sentence subsets, yielding 13.68 hours (${\sim}$16.6k utterances) from 15 speakers: 3 female dysarthric, 3 female healthy controls (HC), 5 male dysarthric, and 4 male HC. The subset comprises 8.66 hours of male and 5.02 hours of female speech, with 7.85 hours of single-word and 5.83 hours of multi-word utterances.

\textbf{NeuroVoz} \cite{mendes2024neurovoz} is a Spanish corpus of Parkinson's disease speech from which sentence repetitions and short monologues were retained, yielding 2.31 hours (1.8k samples) from 111 speakers: 26 female and 28 male HCs, and
20 female and 33 male PD speakers, with 4 of unknown sex. Duration totals 1.28 hours (male), 0.95 hours (female), and the remainder from unknown-sex speakers.

Both datasets are partitioned into train/validation/test splits at a 70/10/20 ratio, stratified by speaker identity to prevent data leakage, with both sexes and pathological and healthy speech represented in every split.

As NeuroVoz is a smaller dataset, we use Spanish Common Voice v24.0 \cite{ardila2020common} to balance Spanish in the training data. We filter 7.3 hours (${\sim}$4.9k samples) of read speech from the original training set, split into train/validation subsets stratified by speaker.

\subsection{MCQA test}
\label{sec:qa-benchmark}

The MCQA test is derived from the metadata of the used databases. NeuroVoz includes sex and age metadata, while TORGO includes sex information only. For each utterance in the test splits, we pose the following questions:

\begin{itemize}
  \item \textbf{Sex} (\textit{``What is the sex of the speaker?''}):
        a three-choice question with options \textit{``Male''},
        \textit{``Female''}, and \textit{``I do not know''}.

  \item \textbf{Age range} (\textit{``What is the age range of the speaker?''}):
        a four-choice question with options \textit{``Less than 20 years old''},
        \textit{``Between 20 and 40 years old''},
        \textit{``Between 40 and 60 years old''}, and
        \textit{``More than 60 years old''}. This question is asked only for
        NeuroVoz.
\end{itemize}

Answer choices are shuffled independently per question to avoid order bias.
The resulting benchmark comprises 5,890 audio-question pairs: 5,725 sex
questions (5,560 from TORGO, 165 from NeuroVoz) and 165 age-range questions
(NeuroVoz only). Both tasks exhibit class imbalance reflecting the demographic
distribution of the test splits: \textit{``Female''} accounts for 59.5\% of sex
questions, and \textit{``More than 60 years old''} for 69.7\% of age-range
questions.

\subsection{Training}
\label{sec:training}

We evaluate five adaptation strategies applied to Voxtral-Mini: \textbf{FFT}, full fine-tuning of all model weights; \textbf{F-LoRA}, LoRA applied to all attention and feed-forward layers; \textbf{EFT}, fine-tuning of the audio encoder and MLP connector only; \textbf{E-LoRA}, LoRA restricted to the encoder and MLP connector; and \textbf{speaker conditioning}, our proposed FiLM-based method that trains the FiLM generator and speaker embedding extractor (Section~\ref{sec:film-conditioning}). For ASR fine-tuning, we use the special \texttt{<repeat>} token reserved for the ASR task in the Voxtral-Mini model.

Hyperparameters were tailored per strategy. FFT used conservative settings
(lr $= 10^{-5}$, 3 epochs) to mitigate catastrophic forgetting; EFT used moderate settings (lr $= 2 \times 10^{-5}$, 5 epochs); LoRA-based methods used more aggressive settings (lr $= 3 \times 10^{-4}$, 10 epochs) with rank
$r = 8$, scaling $\alpha = 16$, and dropout $= 0.1$.

For speaker conditioning, only ${\sim}73.5$M of 4.75B parameters
(${\sim}1.6\%$) are updated: the FiLM bank (${\sim}48.3$M parameters,
lr $= 2 \times 10^{-4}$) and the SiAmResNet34 x-vector extractor
(${\sim}25.2$M parameters, lr $= 2 \times 10^{-5}$). The x-vector extractor is
fine-tuned at a reduced learning rate to adapt toward pathology-relevant features without forgetting speaker representations. Input waveforms are truncated to 15\,s for x-vector extraction to avoid out-of-memory issues. The training is done for 10 epochs.

The FiLM generators are initialized to behave as identity transforms, so
training starts from unmodified base model behavior. Concretely, the output
biases are set to $\boldsymbol{\gamma}_\ell = \mathbf{1}$ and
$\boldsymbol{\beta}_\ell = \mathbf{0}$, and the final linear layer weights are
sampled from $\mathcal{N}(0, 10^{-3})$, small enough to avoid perturbing the
encoder at initialization while keeping gradients flowing. The gate bias is
fixed at $-2$, so $\alpha_\ell = \sigma(-2) \approx 0.12$, meaning the gate
starts nearly closed and opens progressively during training. This ensures
$\widetilde{\mathbf{H}}_\ell \approx \mathbf{H}_\ell$ at step zero
(Eq.~\ref{eq:film-residual}). Combined with the normative masking in
Eq.~\ref{eq:normative-mask}, healthy speech always drives a zero vector into
the generators, so FiLM strictly acts as an identity for normative utterances
throughout training by design.

All strategies share an effective batch size of 16 via gradient accumulation, a cosine learning rate schedule with 100 warm-up steps, and early stopping (patience $= 5$) based on validation loss. The process is optimized with AdamW. All experiments were run on two A100-SXM4-40\,GB GPUs.

\subsection{Evaluation}
\label{sec:eval}
ASR performance is measured by Word Error Rate (WER) on the TORGO and NeuroVoz test splits. For TORGO, we report overall WER alongside single-word and multi-word breakdowns. Overall WER is computed by aggregating errors across all utterances, reflecting the dataset's sample distribution rather than the mean of subgroup WERs. References and hypotheses are normalized by lowercasing, removing punctuation, and converting standalone digits to their word forms.

The ability to answer questions about speech is assessed via MCQA accuracy. Each question is presented as a prompt containing the audio and labeled choices: 
\begin{quote}
    \textit{\{question\} Choose the correct option: (A) \{choice\_0\} (B) \{choice\_1\} (C) \{choice\_2\} (D) \{choice\_3\}}
\end{quote}

The model is expected to respond with the full text of the correct option
(e.g., \textit{``Male''}) or with a leading choice letter (e.g., \textit{``(A) Male''}). If the model outputs only a choice letter (e.g., \textit{``(A)''}), the corresponding choice text is used instead. Accuracy is computed via exact match after lowercasing and removing any choice-letter prefix.

All inferences are obtained via greedy decoding to ensure deterministic and reproducible results across all evaluated systems.

\subsection{Post-processing}
\label{sec:postprocessing}

We evaluate rule-based post-processing (PP) to mitigate ASR hallucinations,
following \cite{tan2025cba}. Three sequential steps are applied: (i) words
exceeding 15 characters is analyzed for internal character-level repetition
and reduced to their base unit; (ii) consecutive repeated words are collapsed
(word-level deduplication); (iii) consecutive repeated phrases are reduced to
a single occurrence (phrase-level deduplication).

\begin{table*}[h]
\centering
\caption{WER (\%) of Voxtral-Mini and adapted variants on the NeuroVoz and
         TORGO test sets. PP = post-processing.}
\vspace{-0.2cm}
\label{tab:asr_results}
\resizebox{1.0\textwidth}{!}{%
\begin{tabular}{lccccccccc}
\toprule
\multirow{3}{*}{\textbf{Model}} &
\multirow{3}{*}{\textbf{Trained blocks}} &
\multicolumn{2}{c}{\textbf{NeuroVoz}} &
\multicolumn{6}{c}{\textbf{TORGO}} \\
\cmidrule(lr){3-4} \cmidrule(lr){5-10}
 & & \multirow{2}{*}{Raw} & \multirow{2}{*}{+PP} &
   \multicolumn{2}{c}{\textbf{Overall}} &
   \multicolumn{2}{c}{\textbf{Single-word}} &
   \multicolumn{2}{c}{\textbf{Multi-word}} \\
\cmidrule(lr){5-6} \cmidrule(lr){7-8} \cmidrule(lr){9-10}
 & & & & Raw & +PP & Raw & +PP & Raw & +PP \\
\midrule
Base         & None
             & 6.75 & 6.87 & 25.15 & 22.09 & 46.83 & 46.33 & 16.13 & 12.00 \\
\midrule
FFT          & Encoder, Connector, Decoder
             & 4.32 & 4.26
             & \textbf{10.97} & \textbf{10.90}
             & \textbf{19.07} & \textbf{19.07}
             & \textbf{7.60}  & \textbf{7.51}  \\
F-LoRA       & Encoder, Connector, Decoder
             & \textbf{4.07} & \textbf{4.13}
             & 12.71 & 12.69 & 22.55 & 22.55 & 8.61 & 8.59 \\
EFT          & Encoder, Connector
             & 4.86 & 4.80 & 19.99 & 14.12 & 28.18 & 23.15 & 16.58 & 10.36 \\
E-LoRA       & Encoder, Connector
             & 5.84 & 5.71 & 18.94 & 16.94 & 31.13 & 30.50 & 13.86 & 11.30 \\
Spk-Cond
             & SiAmResNet34 + FiLM generators
             & 6.57 & 6.63 & 23.24 & 16.36 & 32.14 & 27.09 & 19.54 & 11.89 \\
\bottomrule
\end{tabular}%
}
\end{table*}

\section{Results}

Table~\ref{tab:asr_results} presents the ASR results, while Table~\ref{tab:qa-baselines} reports MCQA accuracy for paralinguistic questions.

\subsection{ASR}
\label{sec:results-asr}

Fine-tuning substantially reduces WER on TORGO, with FFT and F-LoRA cutting overall error roughly in half relative to the base model. Performance on NeuroVoz improves slightly across all methods. Among fully fine-tuned models, F-LoRA achieves the best NeuroVoz WER (4.07\%) but is outperformed by FFT on TORGO (12.71\% vs.\ 10.97\%), suggesting a generalization-adaptation trade-off between parameter-efficient and full fine-tuning.

Speaker-conditioned FiLM achieves only marginal raw improvement on TORGO over the base model (23.24\% vs.\ 25.15\%); however, post-processing closes much of this gap, yielding 16.36\% WER. This pattern suggests that the model captures the acoustic content of pathological speech reasonably well, but produces noisier hypotheses that benefit substantially from rule-based correction. This contrasts with fully fine-tuned models, which produce cleaner outputs and benefit less from post-processing. 

EFT fine-tuning exhibits the largest absolute post-processing gain on TORGO (from 19.99\% to 14.12\%), suggesting that while the encoder learns stronger acoustic representations, the frozen decoder struggles to decode them cleanly without additional correction.

\subsection{Speech MCQA}
\label{sec:results-mcqa}

\begin{table}[h]
  \centering
  \caption{Accuracy (\%) on the MCQA questions for Voxtral-mini and adapted versions.}
  \vspace{-0.2cm}
  \label{tab:qa-baselines}
  \begin{tabular}{lrrr}
    \toprule
    Model & Overall (\%) & Sex (\%) & Age (\%) \\
    \midrule
    Random              & 33.1 & 33.3 & 25.0 \\
    Majority class      & ---  & 59.5 & 69.7 \\
    \midrule
    Base                & 52.7 & 53.5 & 24.8 \\
    FFT                 & 60.9 & 62.0 & 21.8 \\
    EFT                 & \textbf{63.5} & \textbf{64.9} & 16.4 \\
    F-LoRA              &  8.4 &  8.6 &  1.2 \\
    E-LoRA              & 49.2 & 49.8 & \textbf{26.7} \\
    Spk-Cond     & 59.3 & 60.7 & 11.5 \\
    \bottomrule
  \end{tabular}
\end{table}

Table~\ref{tab:qa-baselines} reports MCQA accuracy alongside a random-choice baseline and a majority-class baseline. The base model yields near-random performance on age-range prediction (24.8\%), indicating that it cannot reliably infer speaker age from speech alone. We therefore exclude age prediction from our primary analysis and focus on sex questions, where the base model already exceeds the random baseline (53.5\%) but remains below the majority-class ceiling (59.5\%).

Adaptation strategies differ markedly in their effect on speech MCQA. Notably, both FFT and EFT improve sex accuracy over the base model (62.0\% and 64.9\%, respectively), suggesting that adapting to pathological speech inadvertently sharpens the encoder's sensitivity to speaker-level acoustic features. EFT achieves the highest sex accuracy overall, likely because keeping the LLM decoder frozen preserves the instruction-following behavior required to correctly select among labeled choices. F-LoRA collapses its accuracy to 8.4\%, indicating that applying LoRA across all model layers severely degrades generalization. Observing some of its outputs, it just can not follow the given instructions of selecting a choice. Speaker-conditioned Voxtral achieves competitive sex accuracy (60.7\%), trailing EFT by only 4.2 percentage points while updating a lightweight set of parameters and leaving all base model weights intact, demonstrating a favorable efficiency-generalization trade-off. Crucially, the base model behavior preservation is by design: zeroing the speaker embeddings for non-pathological inputs, ensuring that healthy speech performance is never compromised.

\section{Limitations}
In our approach, setting the speaker embeddings to a zero vector causes FiLM to act as an identity function, preserving the base model's performance. However, this requires prior knowledge of whether the input speech is pathological, which may not always be available in practice. Furthermore, we assess the model's ability to answer paralinguistic questions after pathological ASR adaptation. Expanding the MCQA evaluation to broader speech understanding and exploring the robustness of the FiLM conditioning under noisy conditions and for short-duration utterances, represent promising future directions.

\section{Conclusion}
We proposed FiLM conditioning of a frozen SpeechLLM encoder via derived information form speaker x-vectors. It is a lightweight approach to pathological speech adaptation that maintains the base model untouched. Results on TORGO and NeuroVoz show that the method reduces WER on pathological speech, though with a higher dependence on post-processing compared to fully fine-tuned models. On speaker sex MCQA questions, our approach achieves competitive accuracy at a fraction of the trainable parameters. These results suggest that speaker-conditioning via FiLM is a viable parameter-efficient alternative to other adaptation strategies.

\newpage
\section{Acknowledgments}
This project has been partially funded by the European Union’s Horizon 2020 RIA ELOQUENCE project (Grant Agreement No. 101135916). Views and opinions expressed are, however, those of the author(s) only and do not necessarily reflect those of the European Union or European Commission-EU. Neither the European Union nor the granting authority can be held responsible for them.

Santosh Kesiraju is supported by Ministry of Education, Youth and Sports of the Czech Republic (MoE) through the OP JAK project "Linguistics, Artificial Intelligence and Language and Speech Technologies: from Research to Applications"  (ID:CZ.02.01.01/00/23\_020/0008518).

\bibliographystyle{IEEEtran}

\end{document}